\documentclass[pdflatex]{sn-jnl}

\usepackage{hyperref}
\usepackage{cite}

\usepackage{amsmath}
\usepackage{lineno}

\jyear{2021}%

\theoremstyle{thmstyleone}%
\theoremstyle{thmstyletwo}%

\theoremstyle{thmstylethree}%

\begin{document}


\title[Article Title]{Reversible Attack Based on Local Visible Adversarial Perturbation}


\author[1]{\fnm{Li} \sur{Chen}}\email{E20201021@stu.ahu.edu.cn}

\author[1]{\fnm{Shaowei} \sur{Zhu}}\email{zhusw520@gmail.com}

\author[2]{\fnm{Abel} \sur{Andrew}}\email{andrew.abel@strath.ac.uk}

\author*[3]{\fnm{Zhaoxia} \sur{Yin}}\email{Zhaoxia.edu@gmail.com}

\affil[1]{\orgdiv{Anhui Provincial Key Laboratory of Multimodal Cognitive Computation}, \orgname{Anhui University}, 
\orgaddress{ \city{Hefei}, \postcode{230601}, \country{China}}}

\affil[2]{\orgdiv{Computer and Information Sciences}, \orgname{University Of Strathclyde}, 
\orgaddress{ \city{Glasgow}, \country{Scotland}}}

\affil[3]{\orgdiv{School of Communication \& Electronic Engineering}, \orgname{East China Normal University}, 
\orgaddress{ \city{Shanghai}, \postcode{200241}, \country{China}}}


\abstract{Adding perturbations to images can mislead classification models to produce incorrect results. Recently, researchers exploited adversarial perturbations to protect image privacy from retrieval by intelligent models. However, adding adversarial perturbations to images destroys the original data, making images useless in digital forensics and other fields. To prevent illegal or unauthorized access to sensitive image data such as human faces without impeding legitimate users, the use of reversible adversarial attack techniques is increasing.
The original image can be recovered from its reversible adversarial examples. However, existing reversible adversarial attack methods are designed for traditional imperceptible adversarial perturbations and ignore the local visible adversarial perturbation. In this paper, we propose a new method for generating reversible adversarial examples based on local visible adversarial perturbation. The information needed for image recovery is embedded into the area beyond the adversarial patch by the reversible data hiding technique. To reduce image distortion, lossless compression and the B-R-G (blue-red-green) embedding principle are adopted. Experiments on CIFAR-10 and ImageNet datasets show that the proposed method can restore the original images error-free while ensuring good attack performance.}


%




\maketitle

\section{Introduction}\label{sec1}

Deep learning \cite{lecun2015deep} has been widely used in various tasks, such as autonomous driving \cite{shah2021robustness}, face recognition \cite{niu2021scale}, identification \cite{yan2020identifying}, speech recognition \cite{abel2021speaker}, and image classification \cite{9506383}. However, recent research  \cite{szegedy2013intriguing} has identified that well-designed adversarial examples pose a potential threat to the security of deep learning systems. Szegedy et al. \cite{szegedy2013intriguing} first proposed the concept of adversarial attack, which refers to causing the target model to generate wrong results by adding specific noise to the input image. The added noise is called adversarial perturbation, and an image with specific noise is called an adversarial example. Recent works proposed using adversarial perturbations to protect image privacy from retrieval by malicious models \cite{8682225,rajabi2021practicality,shan2020fawkes}. However, the impact of adding perturbations to protected images is irreversible, which means that the original image is destroyed and loses its use-value in other research fields, particularly digital forensics. Furthermore, since the COVID-19 outbreak, telecommuting and video conferencing have been the norm, and image privacy protection has received more attention. You et al. ~\cite{9428115} proposed a reversible face privacy protection framework that consists of an encoder-decoder. The encoder generates the protected image, and the decoder can restore the original image. However, the decoder cannot restore the original image without distortion, limiting its potential for further security applications.
Therefore, the ability to protect the image data from malicious model access and ensure that authorized users restore the original image without distortion has significant research potential. 

\begin{figure}[h]%
\centering
\includegraphics[width=0.9\textwidth]{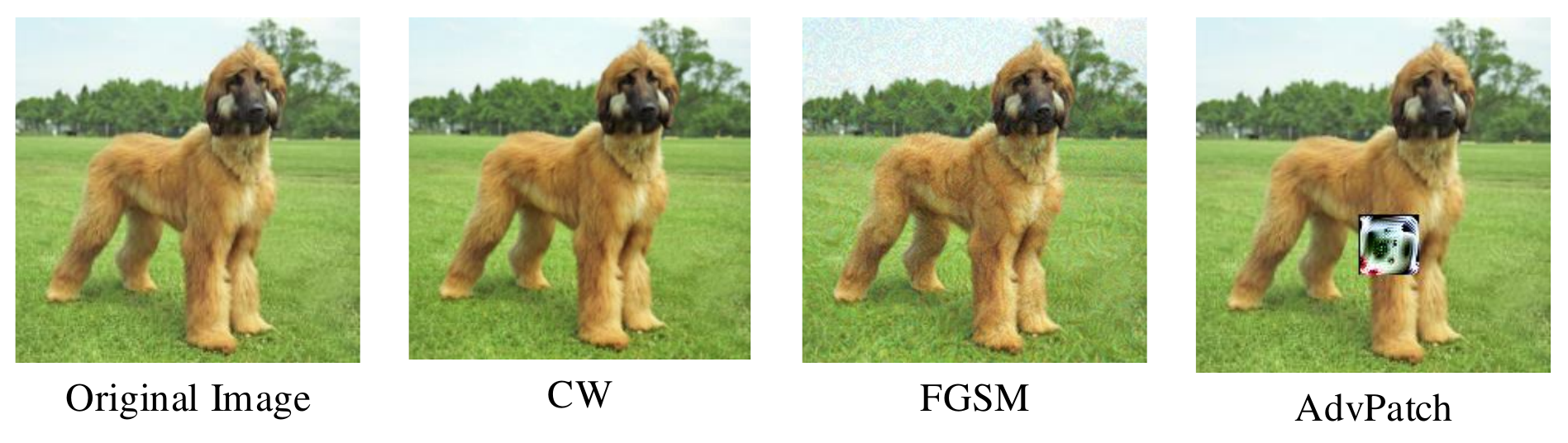}
\caption{Original image and corresponding adversarial examples generated by different attack methods. From left, these are original image, adversarial examples generated by CW \cite{7958570}, FGSM \cite{goodfellow2014explaining} and AdvPatch \cite{brown2017adversarial}}\label{fig1}
\end{figure}

The reversible adversarial attack technology \cite{yin2019reversible} has recently emerged. It generates Reversible Adversarial Examples (RAE) by first using adversarial attack technology to generate adversarial examples and then using reversible information hiding algorithms to embed adversarial perturbations into adversarial examples. On the one hand, RAE \cite{yin2019reversible} can mislead unauthorized models to protect image data. On the other hand, the authorized model can restore the original image without distortion from the reversible adversarial example. However, existing reversible adversarial example generation methods ignore the impact of reversible embedding on attack performance and image visual quality. Furthermore, they aim to generate imperceptible adversarial perturbations and do not consider local visible adversarial perturbation. Figure~\ref{fig1} shows that adversarial examples generated by AdvPatch \cite{brown2017adversarial} have a more significant impact on image content and usability than those generated by imperceptible perturbations, such as Carlini and Wagner (CW) \cite{7958570} and Fast Gradient Sign Method (FGSM) \cite{goodfellow2014explaining}. Thus, reversibility based on local visible adversarial perturbation is even more crucial.

To achieve the reversibility of local visible adversarial perturbation and generate reversible adversarial examples with higher attack performance and better image visual quality, this paper proposes a new reversible attack method based on local visible adversarial perturbation. The contributions of our method are:

(1) The reversibility of local visible adversarial perturbation is explored, and a new reversible attack method based on local visible adversarial perturbation is proposed. 

(2) The WebP compression and the B-R-G embedding principle are adopted to reduce image distortion. Information is embedded in the area beyond the localized adversarial patch to enhance attack performance further. 

(3) Compared with previous methods, the proposed method can successfully generate reversible adversarial examples with better attack performance and image visual quality.

The rest of this paper is organized as follows. Section~\ref{sec-background} provides a detailed literature review, followed by Section~\ref{sec2}, which presents the detailed methodology behind the generation of our proposed reversible adversarial examples. Section~\ref{sec3} presents detailed experiments, evaluation, and analysis. Finally, Section~\ref{sec4} concludes the paper and proposes future research directions.

\section{Background}\label{sec-background}

\subsection{Adversarial Attack}\label{subsec2}
In recent years, adversarial attack has become an important issue. Most works focused on studying imperceptible adversarial perturbations, which can be classified as image-dependent and universal adversarial attacks. Image-dependent attacks \cite{goodfellow2014explaining, 7958570, Moosavi-Dezfooli_2016_CVPR} refer to the adversarial perturbations that must be calculated again for each input image. In contrast, the universal adversarial attacks \cite{Moosavi-Dezfooli_2017_CVPR,8423654} can apply the same adversarial perturbations to any clean images to mislead the target classifier.

In addition to imperceptible adversarial perturbations, Brown et al. \cite{brown2017adversarial} and Karmon et al. \cite{karmon2018lavan} proposed an alternative method to generate adversarial examples, which localized the perturbation to a small region of the image but did not limit the amplitude of the perturbation (meaning that the perturbation could be visibly seen with the eye). The localized perturbation is referred to as an adversarial patch or adversarial sticker. Compared to the imperceptible adversarial perturbations, the adversarial patch has the advantage of being independent of the scene and the input. To generate adversarial examples, Brown et al. \cite{brown2017adversarial} used expectation over transformation \cite{athalye2018synthesizing} to create a noise patch that can be printed and pasted on any images, photographed and presented to the classifier to cause misclassifications. Karmon et al. \cite{karmon2018lavan} showed that a patch generated by modifying only 2\% of the image pixels could attack the most advanced classifiers. They used an optimization-based method and a modified loss function to generate local adversarial perturbation. To improve the visual fidelity and attack performance, Liu et al. \cite{liu2019perceptual} proposed the Perceptual-Sensitive GAN (PS-GAN) framework. A seed patch is converted into an adversarial patch that strongly relates to the attack image through the adversarial process. They also introduced an attention mechanism to predict the critical attack area to enhance the attack ability further. 

\subsection{Reversible Adversarial Attack}\label{subsec3}
Similar to adversarial attack, information hiding is the embedding of secret information by modifying the input signal, which can be divided into digital watermarking \cite{9343885, zhang2022robust}, steganography and steganalysis \cite{1188294, qu2019novel}, and reversible information hiding \cite{8970412, ren2021separable}. Reversible information hiding is of particular interest because it can recover host signals without distortion. The implementation of reversible information hiding can be divided into two main categories: Reversible Image Transformation (RIT) \cite{hou2018reversible} and Reversible Data Hiding (RDH) \cite{1608163}. Reversible image transformation \cite{hou2018reversible} reverses the original image into an arbitrarily selected target image of the same size to generate a camouflaged image almost indistinguishable from the target image. Reversible data hiding \cite{1608163} refers to modifying the image to embed secret data in a fully reversible manner. The secret information can be extracted and the original image restored without distortion at the receiving end. Therefore,
combining information hiding technology and adversarial attack technology has a lot of research potential.

Reversible adversarial examples are generated by embedding adversarial perturbations into adversarial examples using reversible data hiding technology. The original image can be restored without distortion from the reversible adversarial example. To ensure attack performance, adversarial perturbations are strengthened when generating adversarial examples. With the increase of perturbations, the amount of information embedding increases, which leads to the degradation of the attack performance and image visual quality of the generated reversible adversarial examples. Yin et al. \cite{yin2019reversible} combined adversarial attacks and reversible image transformation techniques to generate reversible adversarial examples. The method achieves a reversible transformation between the original image and its adversarial example. It can not only restore the original image without distortion but also is not limited by the perturbation amplitude, providing a higher attack success rate. 

The reversible adversarial attack combines the adversarial attack and reversible information hiding techniques, but there are differences among them. The adversarial attack technique ~\cite{szegedy2013intriguing} can cause the model to generate wrong results by adding adversarial perturbations to the image. But the process is irreversible, and the original image cannot be recovered without distortion from the adversarial example. Reversible data hiding technology ~\cite{1608163} means that the embedded secret information can be extracted, and the original image can also be recovered without distortion. In addition, the image with secret information embedded is not adversarial and will not cause the model to generate wrong results. The reversible adversarial attack ~\cite{yin2019reversible} combines the reversible data hiding technology in the adversarial attack technique. The generated reversible adversarial examples can make the model output wrong results and restore the original image without distortion from the reversible adversarial examples without distortion.

\section{Methodology}\label{sec2}

\begin{figure}[h]%
\centering
\includegraphics[width=1\textwidth]{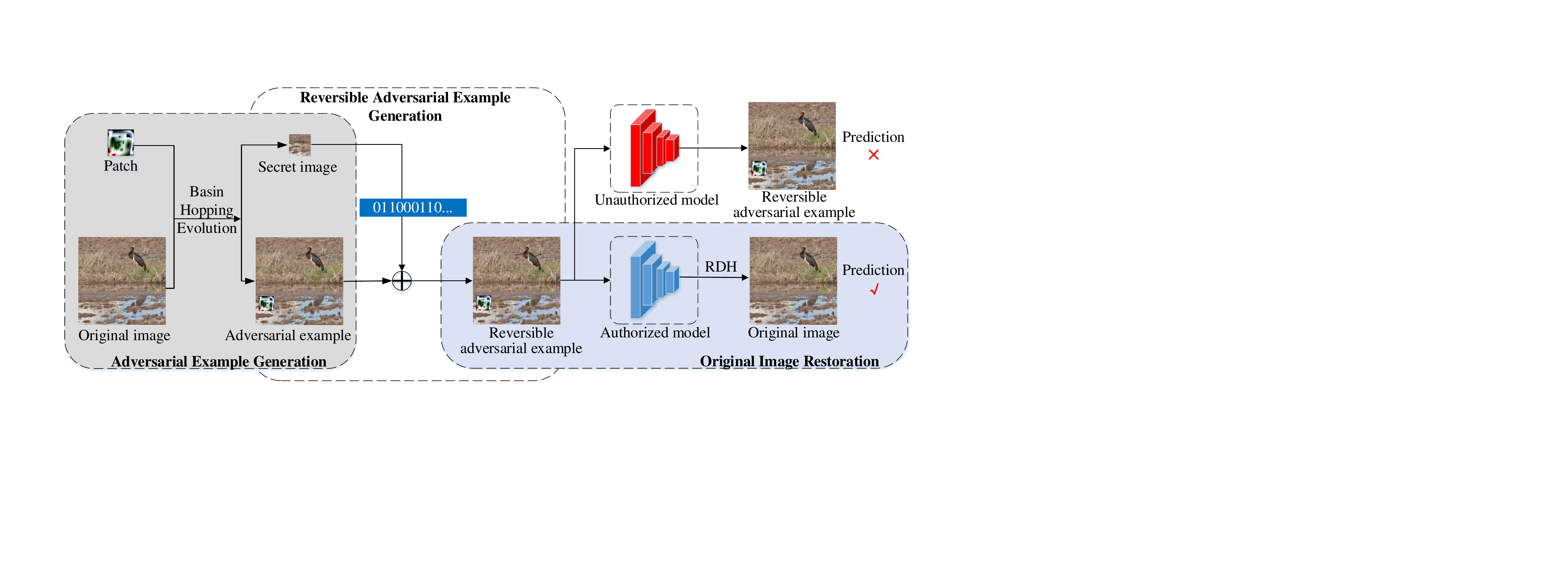}
\caption{The framework of our method. In the adversarial example generation process, the patch location is determined through basin hopping evolution to generate adversarial examples. In the reversible adversarial example generation process, reversible adversarial examples are generated by embedding the information of the part covered by the adversarial patch into adversarial examples. In the original image restoration process, the embedded information is extracted from reversible adversarial examples to restore the original image without distortion}\label{fig2}
\end{figure}

The overall framework is shown in Figure~\ref{fig2}, which consists of adversarial example generation, reversible adversarial example generation, and original image restoration. Next, we will describe our method in detail.

\subsection{Adversarial Example Generation}\label{subsec2}
Adversarial examples must first be generated before reversible adversarial examples can be generated. As discussed previously, there are a number of adversarial example generation methods, but in this paper, AdvPatch \cite{brown2017adversarial} is integrated into our methodology as a suitable adversarial example generation technique for illustrating the generation of reversible adversarial examples. The method can generate a universal adversarial patch applied to any image and has more application value.

Given patch $P$, image $\mathrm{x}$, patch position $l$, patch transformation $t$, and a target class $\hat{y} $, define a patch operation $\mathrm{O}( P,\mathrm{x},l,t)$ which first performs the transformation $t$ on the patch $P$ and then applies the adversarial patch $P$ to the position $l$ of the image $\mathrm{x}$. To obtain the trained patch $\hat{P} $, the attacker uses a variant of the expectation over transformation framework \cite{athalye2018synthesizing} and optimizes the objective function by:
\begin{linenomath*}
\begin{equation}
\widehat{P}=\arg \max _{P} \mathbb{E}_{\mathrm{x} \sim X, t \sim T, l \sim L}[\log \operatorname{Pr}(\hat{y} \mid \mathrm{O}(P, \mathrm{x}, l, t))],\label{eq1}
\end{equation}
\end{linenomath*}
where $X$ denotes the training set, $T$ is a set of transformations including rotations, scaling, etc., and $L$ is a distribution over locations in the image.

Research by Rao et al. \cite{rao2020adversarial} indicates that the patch location within the image plays a critical role in the attacking effectiveness of the image. The Basin Hopping Evolution (BHE) algorithm \cite{jia2020adv} is used to determine the optimal position of the patch in the image. The BHE algorithm \cite{jia2020adv} combines basin hopping and evolutionary algorithms. It has numerous starting points and crossover operations to maintain the diversity of solutions, making it easier to obtain the global optimal solution. Specifically, we first initialize the population and then start the iterative process.  During each iteration, the basin hopping algorithm is used to develop a series of better solutions, followed by crossover and selection operations to select the next generation of the population.

\subsection{ Reversible Adversarial Examples Generation}\label{subsec3}

To achieve reversibility, the part of the original image obscured by
the adversarial patch is taken as the secret image and embedded into the adversarial examples using the RDH algorithm to generate reversible adversarial examples. Next, the generation process of reversible adversarial examples will be described in detail.

To ensure the image visual quality of reversible adversarial examples, the secret image is first compressed by WebP and then converted into binary to reduce the amount of information embedded. In addition, adversarial examples are divided into three channels, R, G, and B. According to the B-R-G principle \cite{9443216}, the same embedding method is used to embed the information in the B, R, and G channels in turn. To further improve attack performance, the data is embedded in the area beyond the adversarial patch. For better understanding, a single channel is used to describe our method. First, the region beyond the adversarial patch is reconstructed into an appropriate size image as a carrier image. Then, we define a flag and a threshold. The flag is used to indicate whether the channel has embedded data. 0 means no embedded data, and 1 means embedded data. The threshold represents the embedded capacity of the channel. Finally, the Prediction Error Extension (PEE) \cite{4099409} algorithm is used for data embedding, which utilizes the correlation of more adjacent pixels with the maximal embedding capacity of 1bpp (bit per pixel). The embedding process can be summarized in the following two steps:

Step 1, computing prediction error. According to the pixel value $a$ and the predicted value $\hat{a}$, the prediction error can be calculated as:
\begin{linenomath*}
\begin{equation}
p = a - \hat{a} .\label{eq2}
\end{equation}
\end{linenomath*}

The predictor predicts the pixel value based on the neighborhood of a given pixel, using the inherent correlation in the pixel neighborhood.  

Step 2, data embedding. The prediction error after embedding a bit $i$ can be calculated as:
\begin{linenomath*}
\begin{equation}
p_{s}=p\oplus  i=2p+i,\label{eq3}
\end{equation}
\end{linenomath*}
where $\oplus$ denotes the difference expansion embedding operation. Then, the pixel value $a_{s}$ is calculated by:
\begin{linenomath*}
\begin{equation}
a_{s} =\hat{a} +p_{s} .\label{eq3}
\end{equation}
\end{linenomath*}

Finally, the patch coordinates, flags, and thresholds are used as auxiliary information, and PEE \cite{4099409} is used to embed the upper left corner of the image. Furthermore, fixed thresholds are used during embedding and passed to the authorized models (i.e., models that are authorized to access the image data).

\subsection{ Original Image Restoration}\label{subsec4}

For unauthorized access to image data, reversible adversarial examples can cause the model to generate wrong results. However, the authorized model can recover the original image without distortion from reversible adversarial examples and correctly access the image data. The image restoration process is divided into three steps:

Step 1, auxiliary information extraction. The auxiliary information is extracted from the upper left of the image, which includes patch coordinates, flags and threshold values. 

Step 2, image data extraction. First, the image is reconstructed according to the patch coordinates. The embedded data is then extracted according to the B-R-G principle \cite{9443216}, based on flags and thresholds.

Step 3, restoration of the original image. The extracted data is saved into WebP format files and then decompressed. Finally, the original image is recovered without distortion according to the patch coordinates. 

\begin{figure}[h]%
\centering
\includegraphics[width=0.9\textwidth]{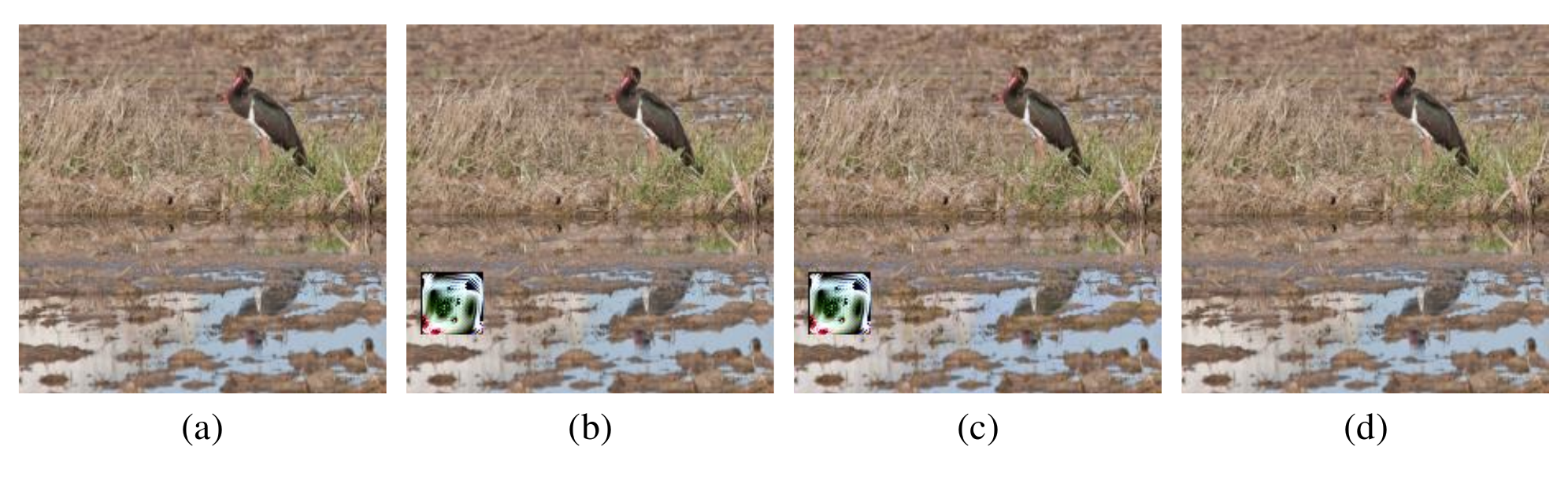}
\caption{Images of the reversible attack process: (a) original image; (b) adversarial example; (c) reversible adversarial example; (d) recovered image}\label{fig3}
\end{figure}

Figure~\ref{fig3} shows an original image and an adversarial example with a localized adversarial patch, a reversible adversarial example, and finally, the restored image resulting from our experiments. 

\section{Experiments and Analysis}\label{sec3}

To illustrate the effectiveness of our proposed method, experiments and analysis of the image classification attack task are performed in this section. 

\subsection{Experiment Settings}\label{setting}

\textbf{Datasets and Models.} 
The ImageNet \cite{russakovsky2015imagenet} and CIFAR-10 \cite{krizhevsky2009learning} datasets are used in experiments. In addition, ResNet50 \cite{He_2016_CVPR}, ResNet101 \cite{He_2016_CVPR}, InceptionV3 \cite{Szegedy_2016_CVPR} and Vgg16 \cite{simonyan2014very} are chosen as threat models. 

\noindent\textbf{Evaluation Metric.} 
We compare our method with Yin et al. \cite{yin2019reversible}. In the following experiments, the method is called RAE\_RIT. The attack success rate, Peak Signal to Noise Ratio (PSNR) \cite{huynh2008scope}, and Structural Similarity (SSIM) \cite{wang2004image} are used as evaluation metrics. The larger value of the attack success rate shows better attack performance, and the larger value of PSNR and SSIM indicates higher image visual quality.

\noindent\textbf{Implement Detail.} 
Adversarial patches with different noise percentages are generated on 2,000 randomly selected images. Noise percentages refer to the adversarial patch size in comparison to the image size. We compare noise percentages of 3\%, 4\%, 5\% and 6\%. After being generated, comparative experiments are conducted on 1,000 randomly selected images that the model could correctly classify. 

\subsection{Comparative Experiments}\label{comparative}

\subsubsection{Attacking Performance}\label{subsubsec2}
Table~\ref{tab1} shows the attacking performance of different methods with different noise percentages. AdvPatch represents the attack success rate of adversarial examples generated by literature \cite{brown2017adversarial}. RAE\_RIT and RAE\_Ours represent the attack success rate of reversible adversarial examples generated by Yin et al. \cite{yin2019reversible} and our method. As can be seen from Table~\ref{tab1}, the attack success rate of our method is higher than RAE\_RIT \cite{yin2019reversible}. The results show that the proposed method can generate reversible adversarial examples with higher attack performance.

\begin{table}[h]
\caption{The attack success rates of adversarial examples and reversible adversarial examples. AdvPatch represents adversarial examples generated by \cite{brown2017adversarial}. RAE\_RIT \cite{yin2019reversible} and RAE\_Ours represent reversible adversarial examples generated by different methods}\label{tab1}%
\centering
\setlength{\tabcolsep}{3.5mm}{
\begin{tabular}{cccccc}
\hline
Dataset                      & Method         & 3\%         & 4\%                 & 5\%         & 6\%                      \\ \hline
\multirow{3}{*}{CIFAR-10}    &   AdvPatch \cite{brown2017adversarial}     &76.32\%                   & 80.60\%          & 97.50\%    &98.10\%                  \\
                             &  RAE\_RIT \cite{yin2019reversible}     &64.20\%                    & 69.62\%          & 92.16\%     & 93.61\%                 \\
                             &  \textbf{RAE\_Ours}     & \textbf{72.73\% }     & \textbf{77.27\% }   & \textbf{95.00\% }         & \textbf{95.32\% }                              \\ \hline
\multirow{3}{*}{ImageNet}   &  AdvPatch \cite{brown2017adversarial}        & 87.70\%                    & 96.60\%            &  99.30\%      & 99.80\%                         \\
                            &  RAE\_RIT \cite{yin2019reversible}    & 73.40\%                  & 85.66\%             & 90.80\%          & 95.30\%               \\
                            &   \textbf{RAE\_Ours}        & \textbf{81.30\% }                  & \textbf{91.60\% }   & \textbf{96.00\% }      &\textbf{98.60\% }           \\ \hline

\end{tabular}}
\end{table}

\subsubsection{Image Visual Quality}\label{subsubsec3}

PSNR is one of the most widely used image similarity evaluation metrics in image processing. The Mean Squared Error (MSE) and PSNR are defined as:
\begin{linenomath*}
\begin{equation}
MSE = \frac{1}{mn}\sum_{x=0}^{m-1}\sum_{y=0}^{n-1} \left [ I(x, y) - J(x, y)\right ] ^{2} ,\label{eq1}
\end{equation}
\begin{equation}
 PSNR=10\times \mathrm{log} _{10}\left ( \frac{MAX_{1}^{2} }{MSE}  \right )  =20\times \mathrm{log} _{10}\left ( \frac{MAX_{1} }{\sqrt{MSE} }  \right )  ,\label{eq2}
\end{equation}
\end{linenomath*}
where $I$, $J$ represents the reference image and test image with size $m\times n$, and $MAX_{1}$ represents the maximum pixel value of the image, which equals 255. The PSNR values of adversarial examples and reversible adversarial examples are shown in Table~\ref{tab2}, and it can be seen that the PSNR value of the RAE\_RIT \cite{yin2019reversible} method remains around 37 and 35 on CIFAR-10 \cite{krizhevsky2009learning} and ImageNet \cite{russakovsky2015imagenet}, respectively, while the PSNR value of our method decreases as the noise percentage increases. However, our proposed method is still better than RAE\_RIT \cite{yin2019reversible} in most cases.

\begin{table}[h]
\caption{Comparison of results of image quality between adversarial examples and reversible adversarial examples with PSNR (dB)}\label{tab2}%
\centering
\setlength{\tabcolsep}{4.5mm}{
\begin{tabular}{cccccc}
\hline
Dataset                      & Method         &3\%         &4\%                 &5\%         &6\%                      \\ \hline
\multirow{2}{*}{CIFAR-10}     &  RAE\_RIT \cite{yin2019reversible}     &37.31                  &37.10          &36.91      &37.03                 \\
                             &  \textbf{RAE\_Ours}     & \textbf{51.23}     & \textbf{50.41}   & \textbf{48.04}         & \textbf{47.40}                              \\ \hline
\multirow{2}{*}{ImageNet}    &  RAE\_RIT \cite{yin2019reversible}    & 34.87                   & 35.30            & 35.19           & \textbf{35.01}                \\
                            &   \textbf{RAE\_Ours}        & \textbf{42.89}                  & \textbf{39.58}                 & \textbf{35.31}      &33.43           \\ \hline

\end{tabular}}
\end{table}

The calculation of PSNR is based on the error between corresponding pixels, while SSIM measures image similarity from luminance $l$, contrast $c$, and structure $s$, respectively.  SSIM is formulated as: 
\begin{linenomath*}
\begin{equation}
SSIM\left ( X, Y \right ) =l\left ( X, Y \right ) \times c\left ( X, Y \right ) \times s\left (  X, Y\right ) ,\label{eq3}
\end{equation}
\end{linenomath*}
 where:
 \begin{linenomath*}
 \begin{equation}
 \left\{\begin{matrix}l\left ( X,Y \right ) =\frac{2\mu _{X}\mu _{Y} +  C_{1}}{\mu _{X}^{2} +  \mu _{Y}^{2} +  C_{1} } 
 \\c\left ( X,Y \right )=\frac{2\sigma _{X}\sigma _{Y} +  C_{2}}{\sigma _{X}^{2} +  \sigma _{Y}^{2} +  C_{2} }
 \\s\left ( X,Y \right )=\frac{\sigma _{XY} +  C_{3} }{\sigma _{X}\sigma _{Y}+  C_{3} } ,\label{eq4}
\end{matrix}\right.
\end{equation}
\end{linenomath*}
 where $\mu _{X}$ and $\sigma _{X}^{2}$ are the mean and variance respectively of image $X$. $\sigma _{XY}$ is the covariance between $X$ and $Y$. $C_{1}$, $C_{2}$ and $C_{3}$ are positive constants to avoid a null denominator. The SSIM values of different methods are shown in  Table~\ref{tab3}.  It is clear that compared with other methods, our method can achieve a higher SSIM value. Furthermore, the SSIM value of our method even reaches 0.99 on CIFAR-10 \cite{krizhevsky2009learning}. Tables~\ref{tab2} and \ref{tab3} demonstrate that the RAEs generated by our method have better image visual quality.
 
\begin{table}[h]
\caption{Comparison results of image quality between adversarial examples and reversible adversarial examples with SSIM}\label{tab3}%
\centering
\setlength{\tabcolsep}{5mm}{
\begin{tabular}{cccccc}
\hline
Dataset                      & Method         & 3\%         & 4\%                 & 5\%         & 6\%                      \\ \hline
\multirow{2}{*}{CIFAR-10}     &  RAE\_RIT \cite{yin2019reversible}     &0.95                   & 0.95         & 0.95     & 0.95                 \\
                             &  \textbf{RAE\_Ours}     & \textbf{0.99}     & \textbf{0.99}   & \textbf{0.99}         & \textbf{0.99}                              \\ \hline
\multirow{2}{*}{ImageNet}    &  RAE\_RIT \cite{yin2019reversible}    & 0.95                  & 0.95            & 0.95           & 0.95               \\
                            &   \textbf{RAE\_Ours}        & \textbf{0.99}                  & \textbf{0.99}                 & \textbf{0.97}      &\textbf{0.95}         \\ \hline

\end{tabular}}
\end{table}

\subsection{Transferability}\label{subsec4}
Because the black-box attack has more practical application value, the transferability of RAEs is tested in the black-box settings. Taking the 4\% noise percentage as an example, RAEs are generated for a source attacked model, then attack all other target models. Table~\ref{tab4} shows the attack success rates of RAEs when transferring attacks between different classification models. It can be seen from Table~\ref{tab4} that RAEs generated by ResNet50 show good attacking performance on Vgg16, but the attacking performance is poor on ResNet101 and InceptionV3. The transferability of adversarial examples has similar results. This is because attacking ability is highly correlated with the capacity of the learning model to generate the adversarial examples and shows our reversible technique does not influence the transferability of examples.


\begin{table}[h]
\begin{center}
\setlength{\tabcolsep}{5mm}{

\caption{ The attack success rates of reversible adversarial examples transferred between different models when the noise percentage is 4\%}\label{tab4}%
\begin{tabular}{ccccc}
\toprule
  & ResNet50  & ResNet101 & InceptionV3 & Vgg16\\
\midrule
ResNet50  & 91.6\%   & 56.0\%   & 54.6\%  & 76.4\%   \\
ResNet101    & 60.4\%    & 95.4\%   & 49.7\%  & 65.7\%   \\
InceptionV3   & 59.9\%    & 54.1\%   & 94.8\%  & 66.9\%   \\
Vgg16    & 56.3\%    & 45.6\%   & 43.0\%  & 99.2\%   \\
\botrule
\end{tabular}}
\end{center}
\end{table}

\subsection{Computational Complexity}\label{time}

\begin{figure}[h]%
\centering
\includegraphics[width=1.0\textwidth]{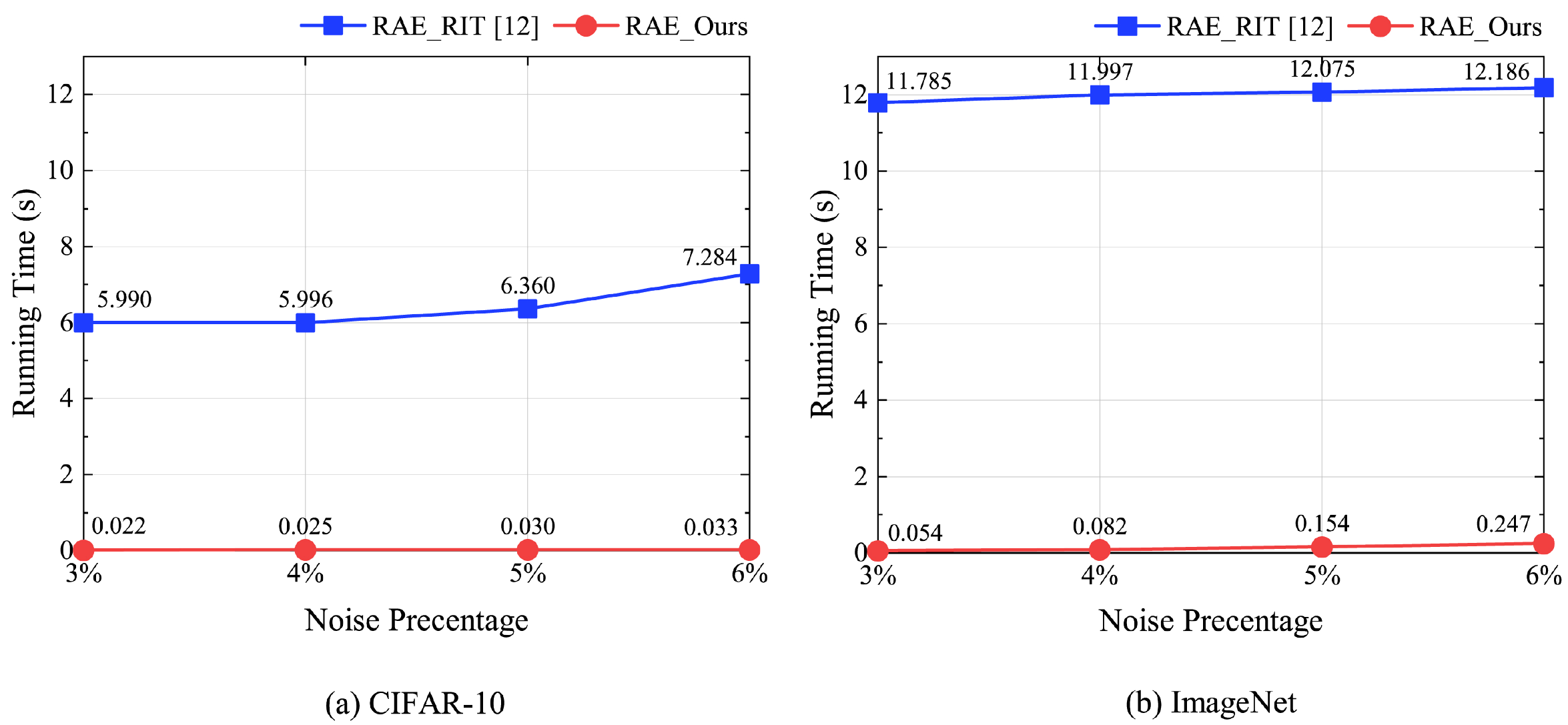}
\caption{Comparison results of computational complexity of reversible embedding process}\label{fig4}
\end{figure}

The running time for generating reversible adversarial examples by different methods has been investigated. The adversarial example generation process is the same. For different noise percentages, the average running time to generate 1,000 adversarial examples on ImageNet \cite{russakovsky2015imagenet} is 8.6s, 2.5s, 0.9s and 0.3s, while the average time on CIFAR-10 \cite{krizhevsky2009learning} is 19.6s, 15.6s, 2.8s, 1.8s. However, the reversible embedding process is different for different methods, and the running time of the reversible embedding process is compared next. The average running time of 1,000 images from CIFAR-10 \cite{krizhevsky2009learning} and ImageNet \cite{russakovsky2015imagenet} datasets is shown in Figure~\ref{fig4}, and it can be seen that the running time of the proposed method is much smaller than RAE\_RIT \cite{yin2019reversible}. Therefore, our method is quicker to use in practice.

\subsection{Results Analysis }\label{analysis}

Compared with previous reversible attack methods, our paper proposed reversible attack based on local visual adversarial perturbation. To guarantee the attack performance of reversible adversarial examples, the information is embedded beyond the patch to preserve the structure of adversarial perturbations better. Experiments also demonstrated the superiority of our method in attack performance. For different noise percentages, the PSNR and SSIM values of RAE\_RIT \cite{yin2019reversible} have smaller fluctuation ranges because the information embedding amount of the method is essentially constant, so even if the noise percentage is small, the image distortion is still severe. In contrast, our method first compresses the embedded information and then adopts the B-R-G embedding principle \cite{9443216} to reduce the impact on image visual quality. Therefore, the proposed method has better image visual quality when the noise percentage is small. With the increase in noise percentage, the amount of information embedding increases, and the embeddable area decreases, resulting in a decrease in image visual quality. However, the proposed method is still better than RAE\_RIT \cite{yin2019reversible} in most cases. In addition, the PSNR value on the CIFAR-10 dataset \cite{krizhevsky2009learning} is better than the ImageNet dataset \cite{russakovsky2015imagenet} because the information embedding of the CIFAR-10 dataset \cite{krizhevsky2009learning} is much smaller than that of the ImageNet dataset \cite{russakovsky2015imagenet} under the same noise percentage.

\section{Conclusion}\label{sec4}

In this paper, we explored the reversibility of adversarial examples based on local visible adversarial perturbation and proposed a reversible adversarial example generation method by embedding information in the area beyond the patch to preserve adversarial capability and achieve image reversibility. To guarantee the image visual quality of the generated adversarial example images, we have to minimize the amount of the data required to be embedded for original image recovery, thus lossless compression is adopted. Compared with the RAE\_RIT method, our proposed method achieves both complete reversibility and state-of-the-art attack performance. As is widely known, for image blocks of the same size, the smoother the image area, the higher the lossless-compression efficiency, and the smaller the amount of compressed data. Therefore, future research will focus on patching the adversarial patch to the smoothest image regions to enhance performance.

\section*{Statements and Declaration}\label{sec5}
\subsection*{Funding}\label{subsec2}
This work is supported by the National Natural Science Foundation of China (No.62172001).
\subsection*{Competing Interests}\label{subsec3}
The authors have no competing interests to declare that are relevant to the content of this article. 
\subsection*{Data Availability Statements}\label{subsec4}
The data that support the findings of this study are available from the corresponding author upon reasonable request.







\begin{appendices}

\end{appendices}



\bibliographystyle{abbrv}
\makeatletter
\renewcommand\@biblabel[1]{#1.}
\makeatother

\end{document}